\documentclass[letterpaper, 10pt, conference]{ieeeconf}
\IEEEoverridecommandlockouts

\let\labelindent\relax
\usepackage[utf8]{inputenc}
\usepackage[T1]{fontenc}
\usepackage{amsmath,amssymb,amsfonts}
\usepackage{algorithm}
\usepackage{algorithmic}
\usepackage{graphicx}
\usepackage{textcomp}
\usepackage{booktabs}
\usepackage{enumitem}
\usepackage{url}
\usepackage{stfloats}
\usepackage{hyperref}
\usepackage{xcolor}
\usepackage{multirow}
\usepackage{array}
\usepackage{bbm}

\hypersetup{
  pdftitle={VEGA: Electric Vehicle Navigation Agent via Physics-Informed Neural Operator and Proximal Policy Optimization},
  pdfauthor={Hansol Lim, Minhyeok Im, Jonathan Boyack, Jee Won Lee, and Jongseong Brad Choi},
  pdfsubject={IROS 2026 conference submission},
  pdfkeywords={electric vehicle routing, reinforcement learning, physics-informed neural operator, energy-aware navigation}
}

\title{\LARGE \bf VEGA: Electric Vehicle Navigation Agent via Physics-Informed Neural Operator and Proximal Policy Optimization}

\author{Hansol Lim, Minhyeok Im, Jonathan Boyack, Jee Won Lee, and Jongseong Brad Choi%
\thanks{Hansol Lim, Jonathan Boyack, Jee Won Lee, and Jongseong Brad Choi are with the Department of Mechanical Engineering, State University of New York, Stony Brook, NY 11794, USA (e-mail: hansol.lim@stonybrook.edu; jonathan.boyack@stonybrook.edu; jeewon.lee@stonybrook.edu; jongseong.choi@stonybrook.edu). Minhyeok Im is with the Department of Computer Science, State University of New York, Stony Brook, NY 11794, USA (e-mail: minhyeok.im@stonybrook.edu).}%
\thanks{This work supported by the National Research Foundation of Korea (NRF) grant funded by the Korea government (MSIT) (No.\ RS-2022-NR067080 and RS-2025-05515607). (Corresponding author: Jongseong Brad Choi).}%
}

\begin{document}

\maketitle
\thispagestyle{empty}
\pagestyle{empty}

\begin{abstract}
We present VEGA, a vehicle-adaptive energy-aware routing system for electric vehicles (EVs) that integrates physics-informed parameter estimation with RL-based charge-aware path planning.
VEGA consists of two coupled modules:
(1)~a physics-informed neural operator (PINO) that estimates vehicle-specific physical parameters---drag, rolling resistance, mass, motor and regenerative-braking efficiencies, and auxiliary load---from short windows of onboard speed and acceleration data;
(2)~a Proximal Policy Optimization (PPO) agent that navigates a charger-annotated road graph, jointly selecting routes and charging stops under state-of-charge constraints.
The agent is initialized via behavior cloning from an A* teacher and fine-tuned with curriculum-guided PPO on the full U.S.\ highway network with Tesla Supercharger locations.
On a cross-country San Francisco--to--New York route (${\sim}$4{,}860\,km), VEGA produces a feasible 20-stop plan with 56.12\,h total trip time and minimum SoC 11.41\%.
Against the controlled Energy-aware A* baseline, the distance and driving-time gaps are small ($-8.49$\,km and +0.37\,h), while inference is $>$20$\times$ faster.
The learned policy generalizes without retraining to road networks in France and Japan.
\end{abstract}

\section{INTRODUCTION}\label{sec:intro}

Electric vehicles (EVs) are increasingly equipped with powerful onboard computers, advancing the vision of software-defined vehicles~\cite{li2023sdv}.
This capability enables onboard AI systems to optimize charge-aware path planning in real time, tailored to each vehicle's current condition and operating environment.
However, accurate per-vehicle energy prediction remains challenging: battery degradation, tire wear, payload variations, and environmental factors cause consumption to drift from factory specifications~\cite{miri2021ev_energy}.
Most navigation systems rely on static models with fixed parameters, leading to range anxiety, suboptimal charging stops, and unplanned detours~\cite{basso2019ev_energy_survey}.

\begin{figure}[t]
\centering
\includegraphics[width=\columnwidth]{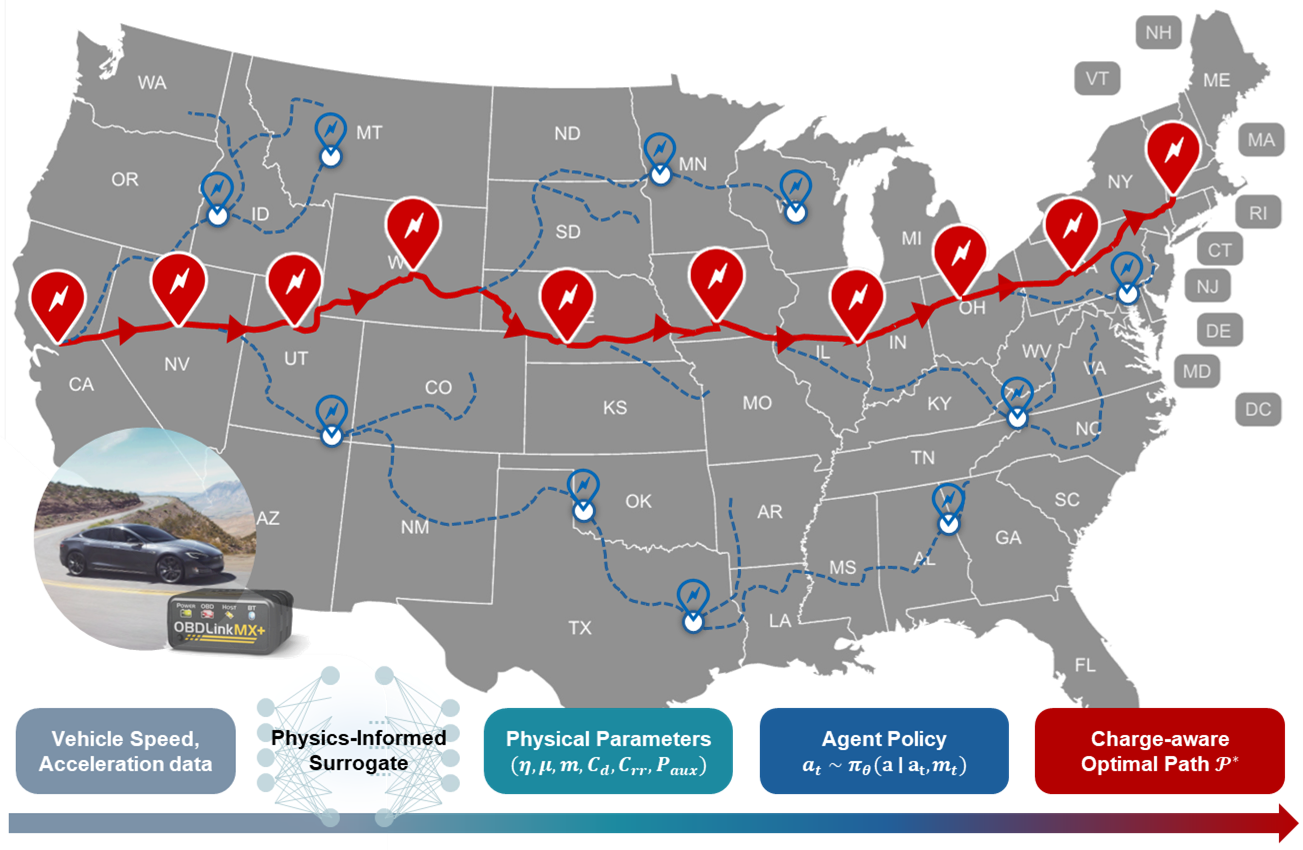}
\caption{VEGA overview. A PINO estimates vehicle-specific parameters from speed and acceleration logs, and a PPO planner uses the resulting physics-based energy model to make charge-aware routing decisions on a charger-annotated road graph.}
\label{fig:vega_overview}
\end{figure}

We present \textbf{VEGA} (EV Energy-aware Guidance Agent), an end-to-end system that integrates \emph{vehicle-adaptive energy modeling} with \emph{scalable charge-aware route planning} on real road networks.
Fig.~\ref{fig:vega_overview} summarizes the end-to-end data flow.
VEGA addresses two coupled challenges simultaneously:
\begin{enumerate}[leftmargin=*]
  \item \textbf{Vehicle-adaptive energy inference}: A physics-informed neural operator (PINO)~\cite{lim2024pino_ev} estimates per-vehicle physical parameters---aerodynamic drag, rolling resistance, mass, motor and regenerative-braking efficiencies, and auxiliary load---from short windows of onboard speed and acceleration logs.
  These parameters feed a physics-based power model, providing accurate, personalized energy consumption estimates without additional sensors.
  \item \textbf{Charge-aware route planning}: A reinforcement learning (RL) agent, trained via Proximal Policy Optimization (PPO)~\cite{schulman2017ppo}, navigates a charger-annotated road graph.
  The agent jointly selects road segments and charging stops while respecting state-of-charge (SoC) constraints, yielding fast planning decisions at inference time.
\end{enumerate}

\noindent\textbf{What VEGA adds beyond existing tools.}
Classical shortest-path planners (e.g., Dijkstra, A*) can incorporate energy costs but scale poorly when SoC constraints require state-space expansion~\cite{basso2019ev_energy_survey}.
Commercial planners (e.g., Tesla Trip Planner) use proprietary models inaccessible for customization or research.
Data-driven routing methods~\cite{nazari2018vrp_rl,kool2019vrp_attention} demonstrate promise on combinatorial problems but typically ignore vehicle-specific energy dynamics.
VEGA uniquely couples (i)~a physics-informed \emph{per-vehicle} parameter inference module with (ii)~a learned charge-aware routing policy on real continental-scale road graphs.
At inference, the policy evaluates a fixed-size action set at each decision step instead of running a full online graph search.

\smallskip\noindent\textbf{Contributions.}
\begin{enumerate}[leftmargin=*]
  \item A \emph{systems integration} of a physics-informed model for EV parameter estimation with a PPO-based charge-aware routing agent, requiring only onboard speed logs from OBD-II data.
  \item Application to real-world Tesla Model~3 Long Range driving data on the full U.S. road graph with Tesla Supercharger locations, demonstrating feasibility on continental-scale networks.
  \item Integration of complementary RL techniques---behavior cloning from an Energy-aware A* teacher, on-policy PPO fine-tuning, and curriculum learning---for scalable charge-aware routing on large real-road graphs.
\end{enumerate}

\section{RELATED WORK}\label{sec:related}

\noindent\textbf{EV energy estimation.}
Physics-based longitudinal models~\cite{fiori2016ev_power} are interpretable but degrade when vehicle parameters drift (aging, payload, tire wear).
Data-driven models~\cite{miri2021ev_energy} can adapt from logs but often sacrifice physical consistency.
Physics-informed models (PINNs/operators)~\cite{raissi2019pinn,li2021fno,lim2024pino_ev,wang2023ev_fno} combine both; VEGA uses this line for vehicle-specific parameter adaptation before routing.

\noindent\textbf{EV routing and charging.}
Classical shortest-path methods (Dijkstra, A*~\cite{hart1968astar}) are efficient for static single-criterion routing, but adding SoC and charging decisions expands the state space and increases search cost~\cite{basso2019ev_energy_survey}.
E-VRP formulations~\cite{schneider2014evrptw,montoya2017evrp} model nonlinear charging and battery feasibility, typically under fixed energy parameters.

\noindent\textbf{Learning-based routing.}
Deep RL and neural combinatorial optimization~\cite{nazari2018vrp_rl,kool2019vrp_attention,chen2022ev_path_planning,joshi2022gnn_routing,kwon2020pomo,bengio2021ml4co,swazinna2021overcoming} provide scalable routing heuristics, but most studies do not couple policy learning with per-vehicle energy adaptation on real continental-scale graphs.
VEGA integrates both: PINO-based energy adaptation and learned charge-aware routing on real road networks.

\section{PROBLEM FORMULATION}\label{sec:problem}

We formalize the charge-aware EV routing problem on a real road network.

\subsection{Road Network and Charger Graph}
Let $\mathcal{G} = (\mathcal{V}, \mathcal{E})$ be a directed road graph extracted from OpenStreetMap, where each node $v \in \mathcal{V}$ stores a geographic coordinate and each edge $(i,j) \in \mathcal{E}$ carries length $d_{ij}$ (km) and speed limit $\bar{v}_{ij}$ (km/h).
A subset $\mathcal{C} \subset \mathcal{V}$ marks charging stations (Tesla Superchargers in our experiments).
Each charger provides a nonlinear charging-time function $T_\text{chg}(b_1 \!\to\! b_2)$ in minutes, described in Sec.~\ref{sec:energy_model}.

\subsection{Objective}
Given a start node $s \in \mathcal{V}$, a goal node $g \in \mathcal{V}$, an initial SoC $b_0 \in [0,100]\%$, and usable battery capacity $B$ (kWh), we optimize over a decision sequence $a_{0:T-1}$.
Each decision is either a move to an outgoing neighbor or a charge action:
\begin{equation}\label{eq:problem_actions}
  a_t \in \mathcal{N}(v_t) \cup \{a_\text{chg}\}.
\end{equation}
The objective is to minimize total trip time until the goal is reached:
\begin{equation}\label{eq:objective}
  \min_{a_{0:T-1}} \sum_{t=0}^{T-1} \Delta t_t
\end{equation}
subject to $v_0=s$, $v_T=g$, and the stepwise dynamics
\begin{equation}\label{eq:problem_transition}
  \begin{aligned}
    v_{t+1} &=
    \begin{cases}
      j, & a_t = j \in \mathcal{N}(v_t),\\
      v_t, & a_t = a_\text{chg},
    \end{cases}\\
    b_{t+1} &=
    \begin{cases}
      b_t - 100\,E_{v_tj}/B, & a_t = j \in \mathcal{N}(v_t),\\
      \min(b_\text{max}, b_t + \Delta b_\text{chg}), & a_t = a_\text{chg},\; v_t \in \mathcal{C}.
    \end{cases}
  \end{aligned}
\end{equation}
with per-step elapsed time
\begin{equation}\label{eq:problem_time}
  \Delta t_t =
  \begin{cases}
    d_{v_tj}/\bar{v}_{v_tj}, & a_t = j \in \mathcal{N}(v_t),\\[1mm]
    T_\text{chg}(b_t \!\to\! b_{t+1})/60, & a_t = a_\text{chg}.
  \end{cases}
\end{equation}
The only hard SoC floor is $b_\text{min}=0\%$:
\begin{align}
  b_t &\geq b_\text{min} \quad \forall\, t \in \{0,\ldots,T\}, \label{eq:soc_constraint} \\
  b_t &\leq b_\text{max} \quad \text{after each charge action}, \label{eq:soc_cap}
\end{align}
where $b_\text{max}{=}80\%$ is the charging cap used throughout the paper.
The 15\% reserve used in the reward is a soft penalty threshold, not an additional hard constraint.

\subsection{Inputs and Assumptions}\label{sec:assumptions}
\begin{itemize}[leftmargin=*]
  \item \textbf{Training inputs}: Vehicle speed and acceleration logs from OBD-II; corresponding battery power measurements; road network from OSM; charger locations.
  \item \textbf{Inference inputs}: Recent speed/acceleration window (${\sim}15$ min) for PINO parameter estimation; start/goal nodes; current SoC.
  \item \textbf{Simplifying assumptions}: (i)~Flat terrain (no road grade); (ii)~Cruising at posted speed limits (no traffic); (iii)~No charger queueing delays; (iv)~Charging curve from reference data.
  These are discussed in Sec.~\ref{sec:discussion}.
\end{itemize}

\subsection{Notation}\label{sec:notation}
Table~\ref{tab:notation} lists the most frequently used symbols; equation-specific symbols are defined locally at first use.

\begin{table}[t]
\caption{Notation Summary}
\label{tab:notation}
\centering
\footnotesize
\renewcommand{\arraystretch}{1.1}
\begin{tabular}{@{}lll@{}}
\toprule
\textbf{Symbol} & \textbf{Description} & \textbf{Unit} \\
\midrule
\multicolumn{3}{@{}l}{\textit{Road graph}} \\
$\mathcal{G}{=}(\mathcal{V},\mathcal{E})$ & Directed road graph & -- \\
$d_{ij}$, $\bar{v}_{ij}$ & Edge length; speed limit & km; km/h \\
$\mathcal{C} \subset \mathcal{V}$ & Set of charging stations & -- \\
\midrule
\multicolumn{3}{@{}l}{\textit{Vehicle parameters (inferred by PINO)}} \\
$\Phi_\psi$ & PINO estimator with weights $\psi$ & -- \\
$\psi$ & PINO network weights & -- \\
$C_d$ & Aerodynamic drag coefficient & -- \\
$C_{rr}$ & Rolling resistance coefficient & -- \\
$m$ & Vehicle mass & kg \\
$A$ & Frontal area & m$^2$ \\
$P_\text{aux}$ & Auxiliary power (HVAC, etc.) & W \\
$\eta$ & Motor/drivetrain efficiency & -- \\
$\mu$ & Regenerative braking efficiency & -- \\
\midrule
\multicolumn{3}{@{}l}{\textit{Energy and battery}} \\
$P_\text{bat}(t)$ & Instantaneous battery power & W \\
$\bar{P}_\text{seg}(v)$ & Segment-level battery power & W \\
$E_{ij}$ & Energy consumed on edge $(i,j)$ & kWh \\
$B$ & Usable battery capacity & kWh \\
$b_t$ & State-of-charge at step $t$ & \% \\
$b_\text{min},\,b_\text{max}$ & SoC lower bound and charging cap & \% \\
$\Delta b_\text{chg}$ & SoC gain per charge action (set to 5) & \% \\
$T_\text{chg}(b_1 \!\to\! b_2)$ & Charging time from SoC $b_1$ to $b_2$ & min \\
\midrule
\multicolumn{3}{@{}l}{\textit{RL agent}} \\
$x_t$ & Observation vector ($\in \mathbb{R}^{26}$) & -- \\
$a_t$ & Action at step $t$ & -- \\
$a_\text{chg}$ & Dedicated charge action index ($=8$) & -- \\
$r_t$ & Reward at step $t$ & -- \\
$\pi_\theta$ & PPO policy with parameters $\theta$ & -- \\
$N_\text{max}$ & Maximum episode length & steps \\
\midrule
\multicolumn{3}{@{}l}{\textit{Physical constants}} \\
$\rho$ & Air density & kg/m$^3$ \\
$g$ & Gravitational acceleration & m/s$^2$ \\
\bottomrule
\end{tabular}
\end{table}

\section{METHODOLOGY}\label{sec:method}

\begin{figure}[t]
\centering
\includegraphics[width=\columnwidth]{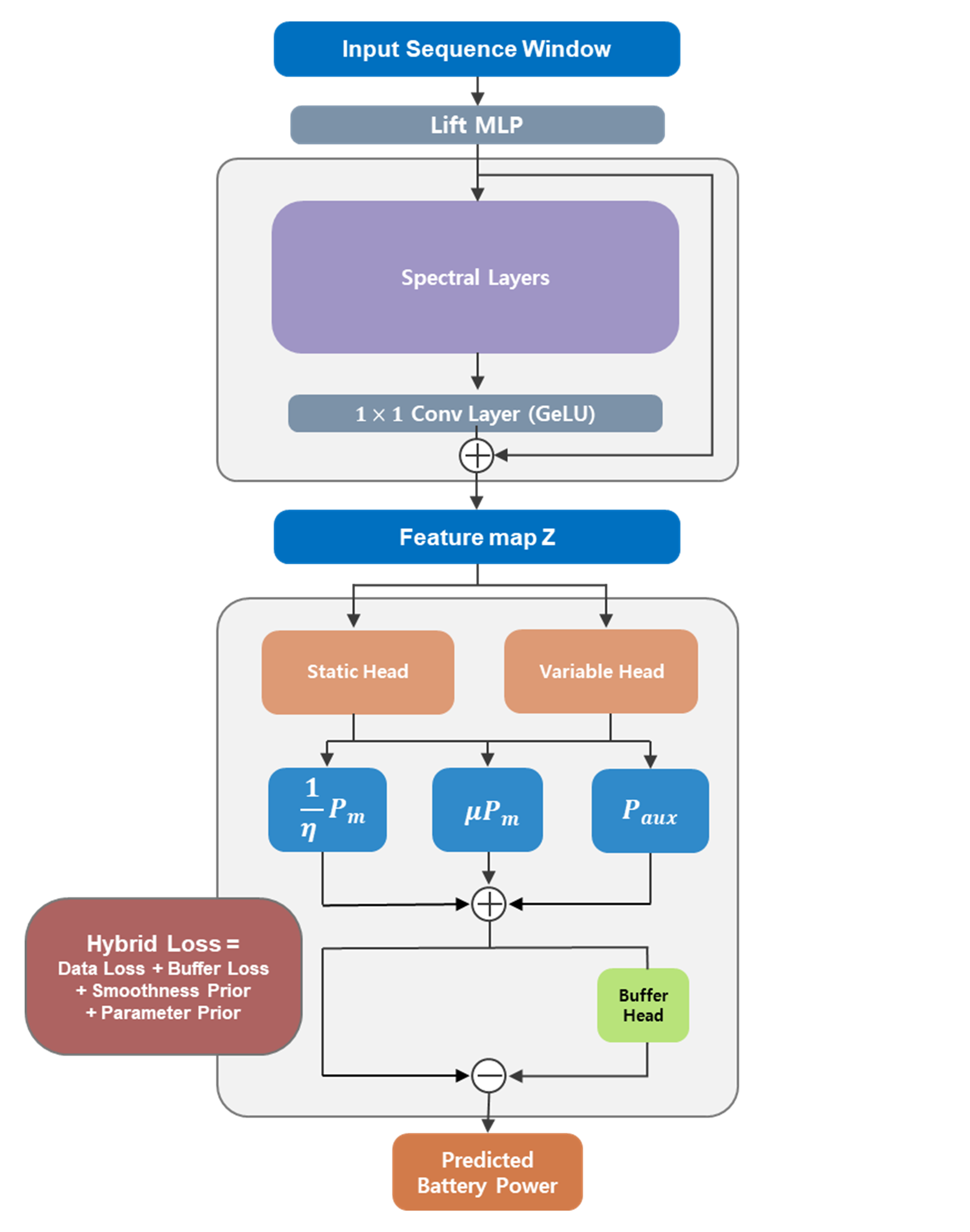}
\caption{VEGA system architecture.  A PINO module estimates vehicle-specific physical parameters from OBD-II speed and acceleration logs.  These parameters feed a physics-based energy model that the PPO routing agent uses to plan charge-aware paths on a real road graph.}
\label{fig:architecture}
\end{figure}

VEGA consists of two coupled modules (Fig.~\ref{fig:architecture}): (i)~a PINO-based parameter estimator that infers vehicle-specific physical constants from speed and acceleration logs, and (ii)~a PPO-trained routing agent that plans charge-aware paths on a real road graph using the estimated parameters.

\subsection{PINO-Based Vehicle Parameter Estimation}\label{sec:pino}

We employ a physics-informed neural operator (PINO)~\cite{lim2024pino_ev} to map a short window of driving data to vehicle-specific parameters.
Let $\Phi_\psi$ denote the neural operator parameterized by weights~$\psi$.
Given a history window $\mathcal{H}$ of speed~$v$ and acceleration~$a$ measurements (approximately 15~minutes of OBD-II data at 10\,Hz), the operator infers six parameters:
\begin{equation}\label{eq:pino_map}
  \Phi_\psi\!: \;\mathcal{H}(v, a) \;\mapsto\; (C_d,\, C_{rr},\, m,\, P_\text{aux},\, \eta,\, \mu).
\end{equation}

The instantaneous battery power is modeled as:
\begin{equation}\label{eq:pbat}
  P_\text{bat}(t) = \frac{1}{\eta}[P_m(t)]_+ - \mu\,[-P_m(t)]_+ + P_\text{aux},
\end{equation}
where $[x]_+ = \max(x,0)$, $P_\text{bat}>0$ denotes battery discharge, and $P_\text{bat}<0$ denotes net charging.
We use the sign convention $P_m>0$ for traction and $P_m<0$ for braking/regen.
Here $\eta \in (0,1]$ is drivetrain efficiency for traction power conversion, and $\mu \in [0,1]$ is the effective regenerative recovery factor.
The mechanical power on flat road is:
\begin{equation}\label{eq:pmech}
  P_m(t) = \tfrac{1}{2}\rho A C_d\, v(t)^3 + C_{rr}\, m g\, v(t) + m\, a(t)\, v(t).
\end{equation}
The first term captures aerodynamic drag, the second rolling resistance, and the third inertial force.
The grade power term ($m g v \sin\alpha$) is omitted under our flat-terrain assumption (Sec.~\ref{sec:assumptions}).
We fix $\rho=1.225$\,kg/m$^3$ and $g=9.81$\,m/s$^2$ throughout.
The frontal area $A$ is treated as a fixed vehicle-geometry constant, while PINO adapts $C_d$ so the drag term $A C_d$ remains vehicle-specific.
The battery capacity $B$ used in SoC updates is the usable capacity (75.0\,kWh for the Tesla Model~3 Long Range in Sec.~\ref{sec:data}).

The operator is trained with a hybrid loss:
\begin{equation}\label{eq:pino_loss}
  \begin{aligned}
    \mathcal{L}_\text{PINO}(\psi) ={}& \frac{1}{N}\sum_{i=1}^{N}\|P_\text{data}^{(i)} - P_\text{pred}^{(i)}\|_F^2 \\
    &+ \lambda_\text{buf}\frac{1}{N}\sum_{i=1}^{N}\|P_\text{res}^{(i)}\|_F^2 + \mathcal{L}_\text{reg},
  \end{aligned}
\end{equation}
where $N$ is the mini-batch size, $\|\cdot\|_F$ is the Frobenius norm, $P_\text{data}^{(i)}$ and $P_\text{pred}^{(i)}$ are measured and predicted battery-power sequences for sample~$i$, $\lambda_\text{buf}$ weights the physics residual term, $P_\text{res}^{(i)}$ is the residual from evaluating~\eqref{eq:pbat}--\eqref{eq:pmech}, and $\mathcal{L}_\text{reg}$ includes smoothness and parameter-range priors (see~\cite{lim2024pino_ev} for details).

\subsection{Segment Energy and Charging Model}\label{sec:energy_model}

For route planning, we compute energy consumption per road segment under a constant-speed approximation.
All physics computations use SI units (m, s, W): road-graph attributes stored as km and km/h are converted to m and m/s before evaluating~\eqref{eq:pseg}--\eqref{eq:energy_seg}.
Between nodes $i$ and $j$ with speed limit $\bar{v}_{ij}$ and segment time $\Delta t_{ij} = d_{ij}/\bar{v}_{ij}$, the segment-level battery power is:
\begin{equation}\label{eq:pseg}
  \bar{P}_\text{seg}(\bar{v}_{ij}) = \frac{1}{\eta}\!\left(\tfrac{1}{2}\rho A C_d\, \bar{v}_{ij}^3 + C_{rr}\,m g\,\bar{v}_{ij}\right) + P_\text{aux}.
\end{equation}
Note that~\eqref{eq:pseg} is a simplified version of~\eqref{eq:pbat}: it drops the acceleration term (constant speed) and the regenerative braking term (no deceleration within a segment), so route-level energy estimates remain conservative in this simplified model.
The segment energy consumption is then:
\begin{equation}\label{eq:energy_seg}
  E_{ij} = \frac{\bar{P}_\text{seg}(\bar{v}_{ij}^{\,\mathrm{SI}})\,\Delta t_{ij}^{\,\mathrm{SI}}}{3.6\times10^6},
\end{equation}
where $E_{ij}$ is in kWh.

\noindent\textbf{SoC update.}
Given battery capacity $B$ and SoC $b_t \in [0,100]\%$, traversing edge $(i,j)$ yields:
\begin{equation}\label{eq:soc_discharge}
  b_{t+1} = b_t - 100\,\frac{E_{ij}}{B}.
\end{equation}
Charging at a station node $v_t \in \mathcal{C}$ updates:
\begin{equation}\label{eq:soc_charge}
  b_{t+1} = \min\!\left(b_\text{max},\; b_t + \Delta b_\text{chg}\right), \qquad \Delta b_\text{chg} = 5.
\end{equation}
Here $b_\text{max}=80\%$ follows Sec.~\ref{sec:problem}. Eq.~\eqref{eq:soc_charge} is applied only when the agent selects the charge action ($a_t{=}a_\text{chg}$) at a charger node ($v_t \in \mathcal{C}$).
The environment masks the charge action when $b_t \ge b_\text{max}$ (equivalently, it sets $q_t{=}0$).

\noindent\textbf{Nonlinear charging time.}
Charging time is nonlinear: charging from 20\% to 80\% takes roughly the same time as 80\% to 100\% (Fig.~\ref{fig:charging_curve}).
We approximate the cumulative charging-time curve in Fig.~\ref{fig:charging_curve} with a 5th-order polynomial.
Define the SoC feature vector $\boldsymbol{\beta}(b) = [b,\,b^2,\,b^3,\,b^4,\,b^5]^\top$ and cumulative charge-time model
\begin{equation}\label{eq:charge_time}
  \begin{aligned}
    T_\text{cum}(b) &= \mathbf{w}^\top \boldsymbol{\beta}(b),\\
    T_\text{chg}(b_1 \!\to\! b_2) &= T_\text{cum}(b_2) - T_\text{cum}(b_1).
  \end{aligned}
\end{equation}
Here $T_\text{cum}$ and $T_\text{chg}$ are measured in minutes; we divide by 60 whenever charging time is added to trip time or search cost.
For each discrete charge action, the elapsed charging time is $\Delta t_t^\text{chg} = T_\text{chg}(b_t \!\to\! b_{t+1})/60$, where $b_{t+1}$ follows~\eqref{eq:soc_charge}.
With the fitted coefficients reported in Sec.~\ref{sec:data}, this gives $T_\text{chg}(20 \!\to\! 80) \approx 32.4$\,min and $T_\text{chg}(80 \!\to\! 100) \approx 33.6$\,min.

\subsection{Charge-Aware Routing as an MDP}\label{sec:mdp}

We formulate the routing problem as a Markov Decision Process (MDP) $(\mathcal{S}, \mathcal{A}, P, R, \gamma)$ on graph~$\mathcal{G}$.

\noindent\textbf{Observation space.}
At each step $t$, the agent observes state $x_t \in \mathbb{R}^{26}$, decomposed into a core state and operational context.
Table~\ref{tab:obs_space} provides the complete specification.

\begin{table}[t]
\caption{Observation Space ($x_t \in \mathbb{R}^{26}$)}
\label{tab:obs_space}
\centering
\scriptsize
\setlength{\tabcolsep}{3pt}
\renewcommand{\arraystretch}{1.05}
\begin{tabular}{@{}p{0.28\linewidth}p{0.44\linewidth}cc@{}}
\toprule
\textbf{Feature} & \textbf{Definition} & \textbf{Type} & \textbf{Dim} \\
\midrule
\multicolumn{4}{@{}l}{\textit{Core state (11 features)}} \\
Current node & Scalar coordinate embedding $\phi(v_t)$ from latitude/longitude (normalized) & Cont. & 1 \\
Goal node & Scalar coordinate embedding $\phi(g)$ from latitude/longitude (normalized) & Cont. & 1 \\
Distance-to-goal $D_t$ & Haversine distance (km) & Cont. & 1 \\
SoC $b_t$ & Battery level (\%) & Cont. & 1 \\
Charger availability & $c(v_t) \in \{0,1\}$ & Bin. & 1 \\
Progress ratio & $1 - D_t / D_0$ & Cont. & 1 \\
Step efficiency & $E_\text{avg}$ per km so far & Cont. & 1 \\
Battery danger: critical & $\mathbb{1}[b_t < 15]$ & Bin. & 1 \\
Battery danger: low & $\mathbb{1}[15 \leq b_t < 25]$ & Bin. & 1 \\
Near-goal flag & $\mathbb{1}[D_t < D_\text{thresh}]$ & Bin. & 1 \\
Near-charger flag & $\mathbb{1}[d_t^\text{chg} < d_\text{chg,thresh}]$ & Bin. & 1 \\
\midrule
\multicolumn{4}{@{}l}{\textit{Operational context (15 features)}} \\
Dist. to nearest charger & $d_t^\text{chg}$ (km) & Cont. & 1 \\
Charge steps remaining $q_t$ & Remaining 5\% steps to $b_\text{max}$ & Cont. & 1 \\
Neighbor distances & $d(u_k, g)$ for $k{=}1{\ldots}8$ & Cont. & 8 \\
Curriculum stage $s$ & Current difficulty level & Cont. & 1 \\
Neighbor count & $|\mathcal{N}(v_t)|$ & Cont. & 1 \\
Episode progress & $t / N_\text{max}$ & Cont. & 1 \\
Start SoC reference & $b^\text{start}$ at charge session & Cont. & 1 \\
SoC margin buffer & $b_t - 15$ (clipped at 0) & Cont. & 1 \\
\bottomrule
\multicolumn{4}{@{}l}{\footnotesize Cont. = continuous, Bin. = binary.}
\end{tabular}
\end{table}
The embedding $\phi(\cdot)$ is graph-independent because it uses normalized coordinates rather than graph-specific node indices (Sec.~\ref{sec:generalization}).
For node $v$ with latitude/longitude $(\ell_v^\text{lat},\ell_v^\text{lon})$, we define
$\phi(v)=\tfrac{1}{2}\!\left(\hat{\ell}_v^\text{lat}+\hat{\ell}_v^\text{lon}\right)$, where
$\hat{\ell}_v^\text{lat}=\frac{\ell_v^\text{lat}-\ell_\text{min}^\text{lat}}{\ell_\text{max}^\text{lat}-\ell_\text{min}^\text{lat}}$ and
$\hat{\ell}_v^\text{lon}=\frac{\ell_v^\text{lon}-\ell_\text{min}^\text{lon}}{\ell_\text{max}^\text{lon}-\ell_\text{min}^\text{lon}}$, with bounds taken from the active road graph.
We define the remaining Table~\ref{tab:obs_space} terms as:
\begin{equation}\label{eq:obs_defs}
  \begin{aligned}
    D_0 &= D_{t=0},\\
    E_\text{avg}(t) &=
    \begin{cases}
      0, & t=0,\\[1mm]
      \dfrac{\sum_{\tau=0}^{t-1} E_{e_\tau}}{\sum_{\tau=0}^{t-1} d_{e_\tau}}, & t>0,
    \end{cases}\\
    q_t &= \left\lceil \frac{\max(0,\,b_\text{max}-b_t)}{\Delta b_\text{chg}} \right\rceil .
  \end{aligned}
\end{equation}
Here $D_0$ is the initial start-goal haversine distance, $e_\tau$ denotes the traversed edge at step $\tau$, $E_\text{avg}$ is in kWh/km, and $q_t=0$ if and only if $b_t \ge b_\text{max}$.
We use fixed thresholds $D_\text{thresh}{=}5$\,km and $d_\text{chg,thresh}{=}3$\,km across all experiments (not tuned per route).

\noindent\textbf{Action space.}
The agent selects from $K{+}1 = 9$ discrete actions.
At each node $v_t$ with neighbors $\mathcal{N}(v_t)$, we preselect $K{=}8$ candidate edges ranked by proximity to the goal (haversine distance from neighbor to $g$).
If $|\mathcal{N}(v_t)| < K$, remaining slots are padded with null actions that repeat the current node.
The padded neighbor-distance features are set to $D_t$ so the observation remains well-defined even when $|\mathcal{N}(v_t)|<K$.
Action $a_t \in \{0, \ldots, 7\}$ selects the $k$-th ranked neighbor.
Action $a_t = a_\text{chg} = 8$ triggers a \emph{charge action}, valid only at supercharger nodes ($c(v_t){=}1$); the charge action is masked when $v_t \notin \mathcal{C}$ or $b_t \ge b_\text{max}$.
Each charge action applies the fixed 5\% SoC increment in~\eqref{eq:soc_charge}, and its duration is $\Delta t_t^\text{chg} = T_\text{chg}(b_t \!\to\! b_{t+1})/60$ from~\eqref{eq:charge_time}.

\noindent\textbf{Feasible action masking.}
We mask the charge action by charger availability and the 80\% SoC cap.
For move actions, we do not assume an additional hard mask by predicted post-edge SoC; infeasibility is instead handled by SoC dynamics, low-SoC penalties, and episode termination when $b_t \le 0$.
An additional deployment-time feasibility mask could compute $b_k^\text{next}=b_t-100E_{v_tu_k}/B$ for each candidate neighbor and suppress actions with $b_k^\text{next}<0$.

\noindent\textbf{Reward.}
The per-step reward combines four components:
\begin{equation}\label{eq:reward}
  r_t = r_t^\text{base} + r_t^\text{bat} + r_t^\text{chg} + r_t^\text{term}.
\end{equation}
\begin{itemize}[leftmargin=*]
  \item $r_t^\text{base}$: step penalty ($-1$) plus progress bonus for reducing distance to goal and a proximity incentive near the destination.
  \item $r_t^\text{bat}$: SoC safety penalty (large negative reward when $b_t < 15\%$; moderate penalty for $15\% \leq b_t < 25\%$); a decision-quality bonus for initiating charging when SoC is low and near a charger.
  \item $r_t^\text{chg}$: penalty for invalid charging attempts; reward for initiating and completing charging sessions, scaled by urgency (lower starting SoC receives larger reward).
  \item $r_t^\text{term}$: success bonus ($+1000$) when $D_t < 5$\,km; battery-depletion penalty ($b_t \leq 0$); timeout penalty ($t \geq N_\text{max}$).
\end{itemize}
The reward coefficients were tuned via iterative experimentation.
The agent is penalized when $b_t < 15\%$ SoC, which corresponds to the buffer that accounts for modeling errors and real-world uncertainty (traffic, temperature, elevation).

\noindent\textbf{Termination.}
An episode terminates when $D_t < 5$\,km (success), $b_t \leq 0$ (battery depletion), or $t \geq N_\text{max}$ (timeout/truncation).

\subsection{Training: Behavior Cloning, PPO, and Curriculum}\label{sec:training}

Training proceeds in two phases, summarized in Algorithm~\ref{alg:training}.

\noindent\textbf{Phase~A: Teacher pretraining.}
We pre-generate expert trajectories using a budgeted A* planner over the augmented search state $z=(v,b)$, where $v \in \mathcal{V}$ and $b$ is discretized in 5\% SoC bins to match the charge-action granularity.
The term \emph{budgeted} refers to this explicit battery budget: successor states that would drive SoC below 0 are pruned during search.
Move successors $(i,b)\!\to\!(j,\,b-100E_{ij}/B)$ use edge cost
\begin{equation}\label{eq:astar_cost}
  c_{ij} = w_t\,T_{ij} + w_e\,E_{ij},
\end{equation}
where $T_{ij} = d_{ij}/\bar{v}_{ij}$ is travel time, $E_{ij}$ is edge energy from~\eqref{eq:energy_seg}, and $w_t, w_e > 0$ are fixed scalar weights.
At charger nodes, the teacher also expands an explicit charge successor $(i,b)\!\to\!(i,\min(b_\text{max}, b+\Delta b_\text{chg}))$ with cost
\begin{equation}\label{eq:astar_charge_cost}
  c_\text{chg}(b) = \frac{w_t}{60}\,T_\text{chg}\!\left(b \!\to\! \min(b_\text{max}, b+\Delta b_\text{chg})\right).
\end{equation}
The heuristic is:
\begin{equation}\label{eq:heuristic}
  h(i) = w_t \frac{d(i,g)}{v_\text{max}} + w_e\,\tilde{E}(i, g),
\end{equation}
where $d(i,g)$ is haversine distance, $v_\text{max}$ is the global speed-limit upper bound, and $\tilde{E}(i,g)$ is the remaining-energy estimate obtained by evaluating~\eqref{eq:energy_seg} on a virtual edge of length $d(i,g)$ at speed $v_\text{max}$.
This heuristic is used as a consistent guide for efficient teacher rollouts; teacher optimality is not required because trajectories are used for behavior cloning pretraining.

The policy $\pi_\theta$ is initialized via behavior cloning on teacher trajectories over short routes:
\begin{equation}\label{eq:bc_loss}
  \mathcal{L}_\text{BC}(\theta) = -\sum_{(s,a) \in \mathcal{D}_\text{teacher}} \log \pi_\theta(a \mid s).
\end{equation}
Here, $\mathcal{D}_\text{teacher}$ denotes the teacher-generated state-action dataset.
In this work, we use Energy-aware A* as the teacher, but the behavior-cloning stage is planner-agnostic: VEGA can be retrained with trajectories from other teachers (e.g., Dijkstra or ant-colony optimization), after which PPO reinforces/adapts those strategies under the policy objective; systematic study of policy behavior under different or mixed teachers is future work.

\noindent\textbf{Phase~B: PPO fine-tuning with curriculum.}
After pretraining, we fine-tune $\pi_\theta$ using PPO~\cite{schulman2017ppo} with generalized advantage estimation~\cite{schulze2020gae}.
During PPO rollouts, the policy acts \emph{autonomously}: no teacher actions override or replace the agent's decisions, ensuring on-policy consistency.
Training uses a short-to-long curriculum~\cite{bengio2020curriculum} with six distance stages:
\begin{equation}\label{eq:curriculum}
  D_s \in \{10, 25, 50, 100, 200, 300\}\text{ km},\quad s = 1, \ldots, 6,
\end{equation}
followed by expansion to long-distance routes up to 3000\,km.
The learning rate and batch size are annealed across stages.

\begin{algorithm}[t]
\caption{VEGA Training Pipeline}
\label{alg:training}
\footnotesize
\begin{algorithmic}[1]
\REQUIRE Road graph $\mathcal{G}$, charger set $\mathcal{C}$, PINO estimator $\Phi_\psi$, inferred params $(C_d, C_{rr}, m, P_\text{aux}, \eta, \mu)$
\STATE \textbf{Phase A: Behavior Cloning}
\FOR{$s = 1$ to $S_\text{short}$}
  \STATE Sample start/goal pair with distance $\sim D_s$
  \STATE Run A* teacher with cost~\eqref{eq:astar_cost} $\to$ trajectory $\tau^\text{teach}$
  \STATE Store $(s_t, a_t^\text{teach})$ pairs in $\mathcal{D}_\text{teacher}$
\ENDFOR
\STATE Train $\pi_\theta$ via $\mathcal{L}_\text{BC}$~\eqref{eq:bc_loss}
\STATE \textbf{Phase B: PPO Fine-tuning}
\FOR{curriculum stage $s = 1$ to $6$}
  \FOR{episode $= 1$ to $M_s$}
    \STATE Sample start/goal with distance $\sim D_s$
    \STATE Rollout $\pi_\theta$ autonomously $\to$ trajectory $\tau$
    \STATE Compute rewards $\{r_t\}$, advantages $\{\hat{A}_t\}$
    \STATE Update $\theta$ via PPO clipped objective
  \ENDFOR
\ENDFOR
\STATE Expand to long routes $D_\text{exp} \in \{D_s, \ldots, 3000\}$\,km
\end{algorithmic}
\end{algorithm}
Here $S_\text{short}$ is the number of short-route behavior-cloning stages, $M_s$ is the number of PPO episodes at curriculum stage $s$, and $D_\text{exp}$ denotes expanded long-route distances.

\section{EXPERIMENTAL EVALUATION}\label{sec:experiments}

\subsection{Environment and Data}\label{sec:data}

\noindent\textbf{Road graph.}
We extract the U.S.\ highway network from OpenStreetMap, yielding a directed graph with ${\sim}$320{,}000 nodes and ${\sim}$730{,}000 edges.
Each edge encodes distance (km), speed limit (km/h), and geographic coordinates.
We overlay 2{,}011 Tesla Supercharger locations as charger nodes ($\mathcal{C}$).

\noindent\textbf{Vehicle data.}
Real-world driving logs are collected from a Tesla Model~3 Long Range with usable battery capacity $B = 75.0$\,kWh via an OBD-II scanner (OBDLink MX+ model) at 10\,Hz, recording speed, acceleration, and battery power.
Battery power is used during PINO training; online deployment uses speed/acceleration only for PINO updates and policy inference.
A 15-minute driving window is sufficient for PINO parameter convergence (Fig.~\ref{fig:pino_convergence}).

\begin{figure}[t]
\centering
\includegraphics[width=0.85\columnwidth]{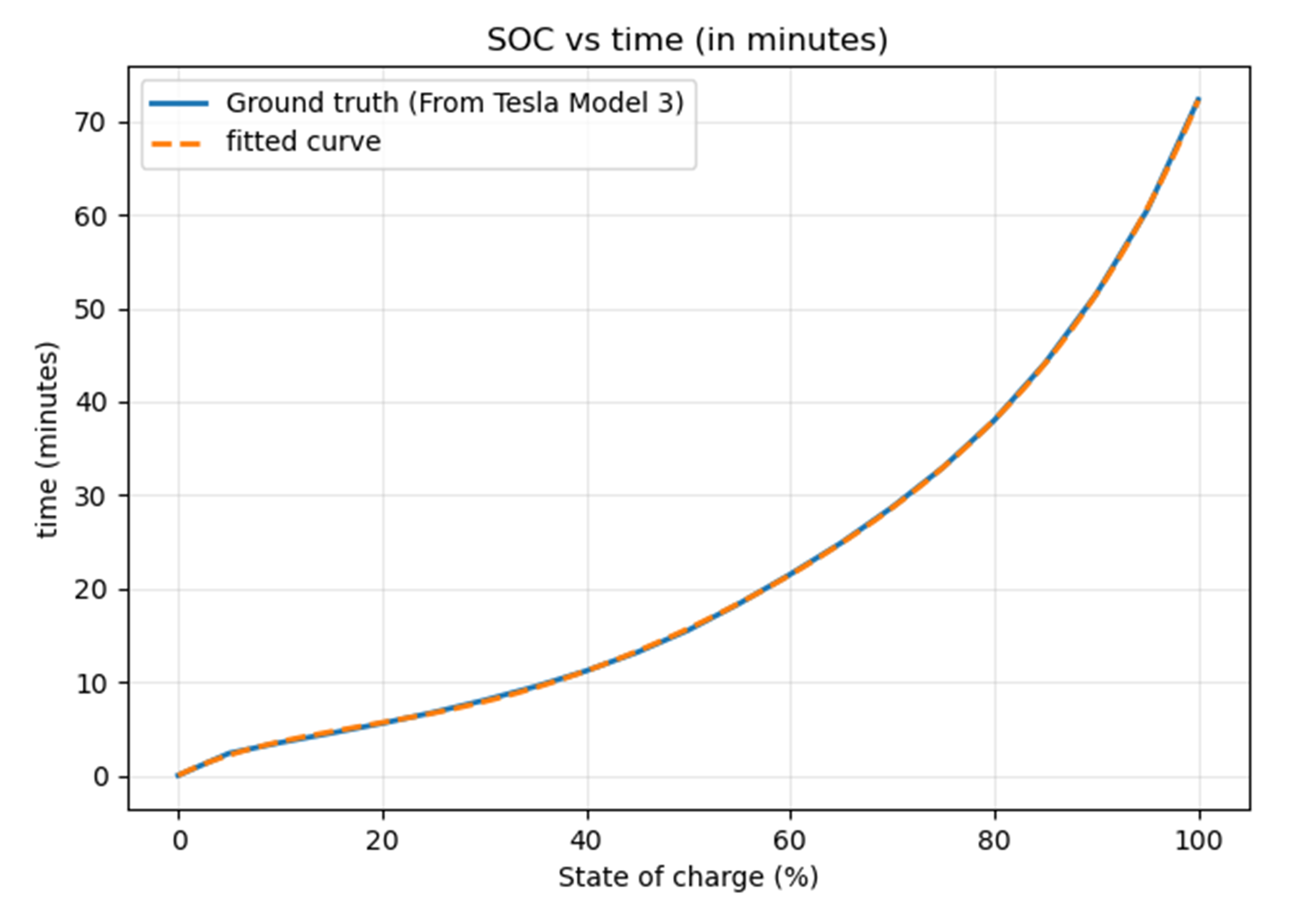}
\caption{Nonlinear charging profile for the Tesla Model~3 Long Range. Charging from 80\% to 100\% SoC takes approximately as long as 20\% to 80\%, motivating the 80\% charging cap used by VEGA.}
\label{fig:charging_curve}
\end{figure}

\noindent\textbf{Charging curve.}
We approximate the benchmark cumulative SoC-vs-time curve in Fig.~\ref{fig:charging_curve} with the polynomial in Eq.~\eqref{eq:charge_time}.
The fitted coefficients are
$w_1{=}5.123{\times}10^{-1}$,
$w_2{=}-1.889{\times}10^{-2}$,
$w_3{=}5.136{\times}10^{-4}$,
$w_4{=}-5.500{\times}10^{-6}$,
and $w_5{=}2.461{\times}10^{-8}$.
These coefficients give $T_\text{chg}(20 \!\to\! 80)\approx32.39$\,min and $T_\text{chg}(80 \!\to\! 100)\approx33.59$\,min, matching the curve shape used by the planner.

\noindent\textbf{Implementation.}
Training and runtime measurements ran on Ubuntu 24.04 LTS with an AMD Ryzen 9 9950X CPU and NVIDIA RTX 5090 (32\,GB), using PyTorch 2.5.1 and CUDA 12.4.
PPO hyperparameters: learning rate $3{\times}10^{-4}$ (annealed), discount $\gamma{=}0.99$, GAE $\lambda{=}0.95$, clip ratio $\epsilon{=}0.2$, batch size 2048 (increased during curriculum).

\subsection{Baselines and Metrics}\label{sec:baselines}

We compare VEGA against:
\begin{enumerate}[leftmargin=*]
  \item \textbf{Tesla Trip Planner}: The commercial reference system accessed through the Tesla vehicle interface.  It uses proprietary models and live data (traffic, weather, charger availability).
  \item \textbf{Energy-aware A*}: Our budgeted A* teacher (Sec.~\ref{sec:training}) run with full SoC tracking on the same road graph and PINO-based segment energy model.
  It is a search-based reference minimizing the time-energy edge cost in Eq.~\eqref{eq:astar_cost}; VEGA instead optimizes PPO return in Eq.~\eqref{eq:reward}, so their objectives are related but not identical.
  \item \textbf{Shortest-time Dijkstra + greedy charging}: Dijkstra's algorithm minimizing travel time only, combined with a reactive charging heuristic---when SoC drops below 20\%, detour to the nearest reachable charger and charge to 80\%.
\end{enumerate}
Tesla comparison is a real-world reference, not a strict direct benchmark, because Tesla uses proprietary energy models and live context.
Our primary quantitative comparisons are the two open baselines (E-A* and Dijkstra+Greedy) under identical simplified assumptions (flat-terrain segment model with haversine distance and speed-limit attributes, no altitude term).
We do not include Google Maps as a baseline because Google Maps do not provide a charger-aware path optimization.

\subsection{PINO Validation}\label{sec:pino_results}

Table~\ref{tab:pino} shows the parameters estimated by PINO from 15~minutes of Tesla Model~3 driving data compared to factory reference specifications.

\begin{figure}[t]
\centering
\includegraphics[width=0.85\columnwidth]{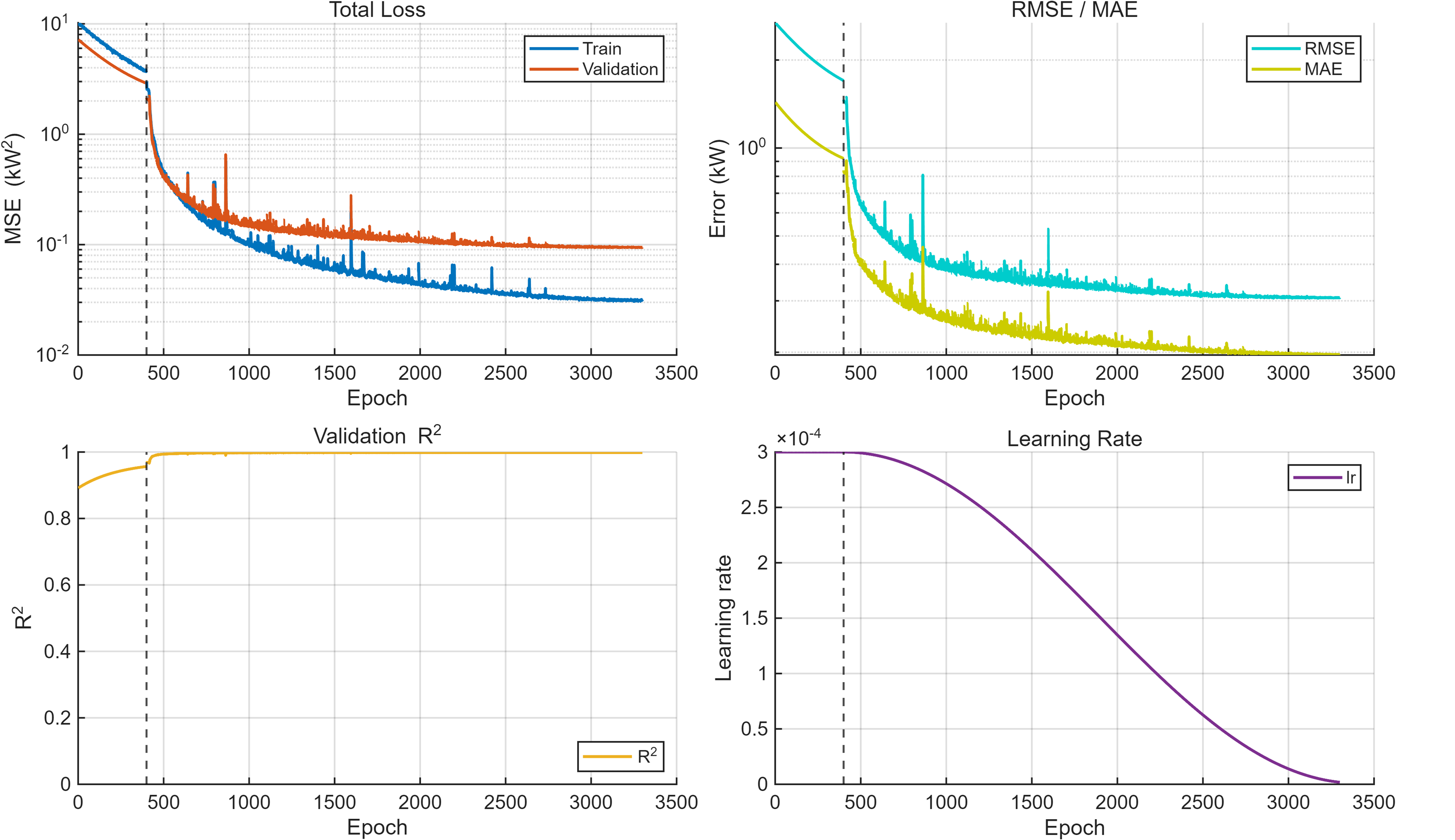}
\caption{Convergence of PINO-estimated parameters over a 15-minute driving window. Estimates stabilize within approximately 10 minutes across multiple trials, demonstrating reliable real-time parameter inference.}
\label{fig:pino_convergence}
\end{figure}

\begin{table}[t]
\caption{PINO-Estimated vs.\ Factory Reference Parameters}
\label{tab:pino}
\centering
\scriptsize
\setlength{\tabcolsep}{3pt}
\begin{tabular}{@{}lcccccc@{}}
\toprule
 & $\eta$ & $\mu$ & $m$\,(kg) & $C_{rr}$ & $C_d$ & $P_\text{aux}$\,(W) \\
\midrule
Factory ref. & -- & -- & 1{,}844 & 0.00960 & 0.23 & -- \\
PINO estimate & 0.830 & 0.741 & 1{,}977 & 0.00946 & 0.235 & 1{,}046 \\
\bottomrule
\end{tabular}
\vspace{1mm}

\footnotesize Factory specifications for $\eta$, $\mu$, and $P_\text{aux}$ are not publicly available; ``--'' indicates no reference value.
The estimated mass (1{,}977\,kg) exceeds the curb weight (1{,}844\,kg), consistent with passenger and cargo loading during data collection.
\end{table}

\noindent The PINO estimates are physically plausible: $C_d{=}0.235$ closely matches the reference 0.23; $C_{rr}{=}0.00946$ aligns with typical tire data; mass is slightly elevated due to payload.
The motor efficiency ($\eta{=}0.83$) and regen coefficient ($\mu{=}0.74$) fall within expected ranges for permanent-magnet synchronous motors.
Fig.~\ref{fig:pino_convergence} shows stable convergence of parameter estimates within the 15-minute window across multiple trials.

\subsection{Controlled Routing Benchmarks}\label{sec:routing_results}

\subsubsection{Cross-Country Route (San Francisco to New York)}

Our main controlled benchmark is the transcontinental SF--NY route (${\sim}$4{,}860\,km), which exceeds the maximum 3{,}000\,km curriculum-training distance by about 1{,}860\,km.
Table~\ref{tab:main_comparison} compares VEGA against Energy-aware A* and Dijkstra+Greedy under the same simplified flat-terrain model; the Tesla column is retained only as a real-world reference for the same start/goal pair.

\begin{table}[t]
\caption{Routing Comparison: San Francisco to New York (${\sim}$4{,}860\,km)}
\label{tab:main_comparison}
\centering
\footnotesize
\renewcommand{\arraystretch}{1.1}
\begin{tabular}{@{}lcccc@{}}
\toprule
\textbf{Metric} & \textbf{VEGA} & \textbf{E-A*} & \textbf{Dijkstra} & \textbf{Tesla} \\
 & & & \textbf{+Greedy} & \textbf{Planner} \\
\midrule
Distance (km) & 4{,}854.96 & 4{,}863.45 & 4{,}831.20 & 4{,}816.77 \\
Driving time (h) & 45.88 & 45.51 & 44.95 & 44.67 \\
Charging time (h) & 10.24 & 8.17 & 9.42 & 9.33 \\
Total trip time (h) & 56.12 & 53.68 & 54.37 & 54.00 \\
Charging stops & 20 & 20 & 24 & 20 \\
Avg. dwell (min) & 30.72 & 24.51 & 23.55 & 27.99 \\
Energy (kWh) & 806.79 & 723.68 & 798.35 & N/A \\
Intensity (Wh/km) & 166.18 & 148.91 & 165.23 & N/A \\
Min SoC (\%) & 11.41 & 15.12 & 8.73 & N/A \\
Planning time (s) & $<2.0$ & $\sim45.0$ & $\sim54.0$ & N/A \\
\bottomrule
\end{tabular}
\vspace{1mm}

\footnotesize E-A* = Energy-aware A* (teacher planner). Tesla Planner values from the vehicle interface; energy/SoC data not available. Dijkstra+Greedy charges reactively when SoC $<$ 20\%. Planning times are route-level wall-clock values on our local Ryzen 9 9950X + RTX 5090.
\end{table}

VEGA and Energy-aware A* both use 20 charging stops, while Dijkstra+Greedy requires 24 because its charger decisions are reactive.
VEGA is already close to E-A* in route geometry: the distance gap is $-8.49$\,km and the driving-time gap is $+0.37$\,h.
\noindent\textbf{Interpreting the optimality gap.}
Relative to E-A*, VEGA shows a $+2.44$\,h total-trip-time gap, split into $+0.37$\,h driving and $+2.07$\,h charging.
The charging gap is exactly matched by the dwell-time difference at the same 20 stops: $(30.72-24.51)$\,min/stop $\times 20=124.2$\,min $=2.07$\,h.
Relative to Dijkstra+Greedy, VEGA reduces the number of charging stops from 24 to 20 while keeping nearly identical energy intensity (166.18 vs.\ 165.23\,Wh/km).
This pattern is consistent with VEGA's design: PPO makes local decisions over a restricted top-$K$ action set, charging is discretized in 5\% steps with an 80\% cap, and the reward favors safe completion over aggressive low-SoC behavior.
VEGA and E-A* share the same PINO parameters and segment-energy model but optimize different objectives (Eq.~\eqref{eq:reward} vs.\ Eq.~\eqref{eq:astar_cost}), so some trip-quality gap is expected.
VEGA therefore provides fast feasible approximate planning ($<2$\,s vs.\ $\sim45$\,s for E-A*), not global optimality.
VEGA remains feasible on SF--NY (minimum SoC 11.41\% $>0$), and the 15\% reserve in the reward is a soft penalty threshold, not a hard constraint.

\begin{figure*}[!b]
\centering
\begin{minipage}[t]{0.32\textwidth}
\centering
\includegraphics[width=\linewidth]{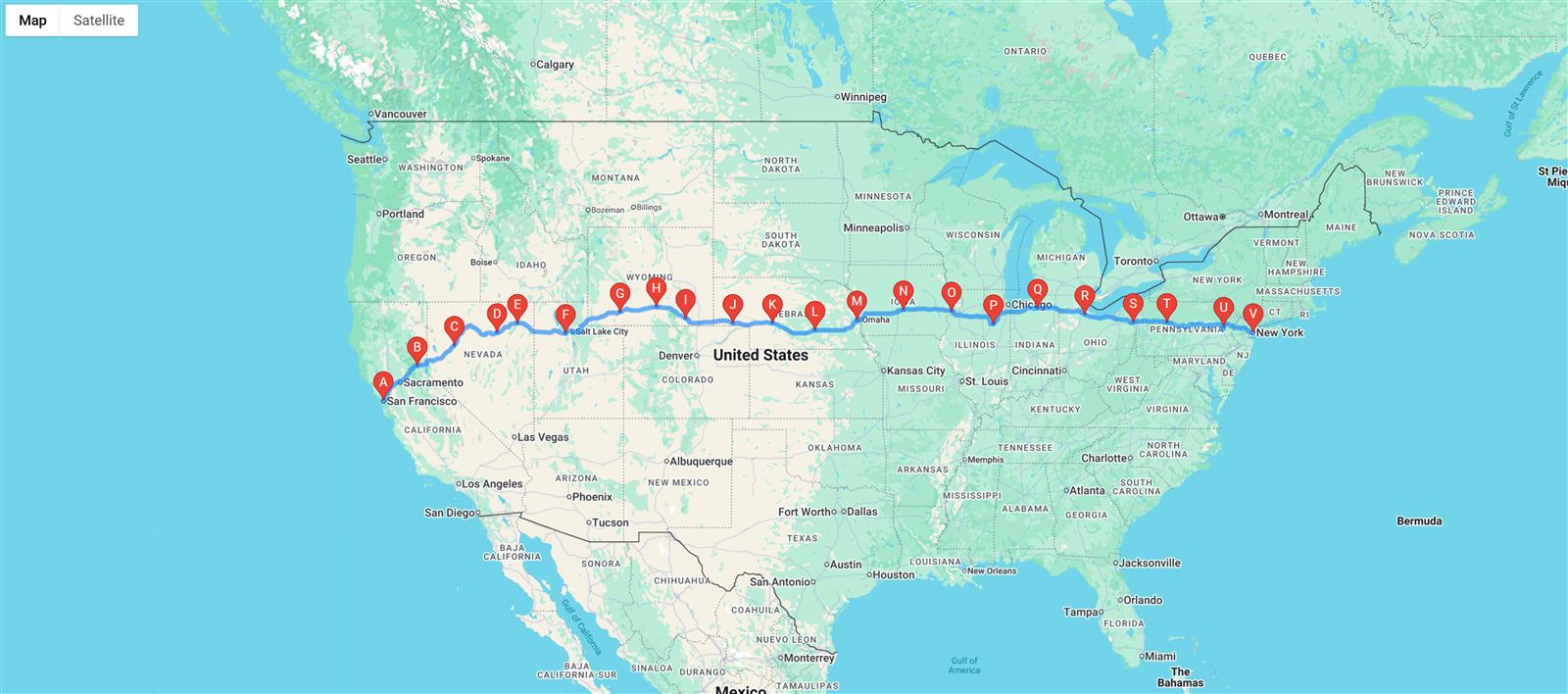}\\[-1mm]
\footnotesize (a) San Francisco--New York
\end{minipage}\hfill
\begin{minipage}[t]{0.32\textwidth}
\centering
\includegraphics[width=\linewidth]{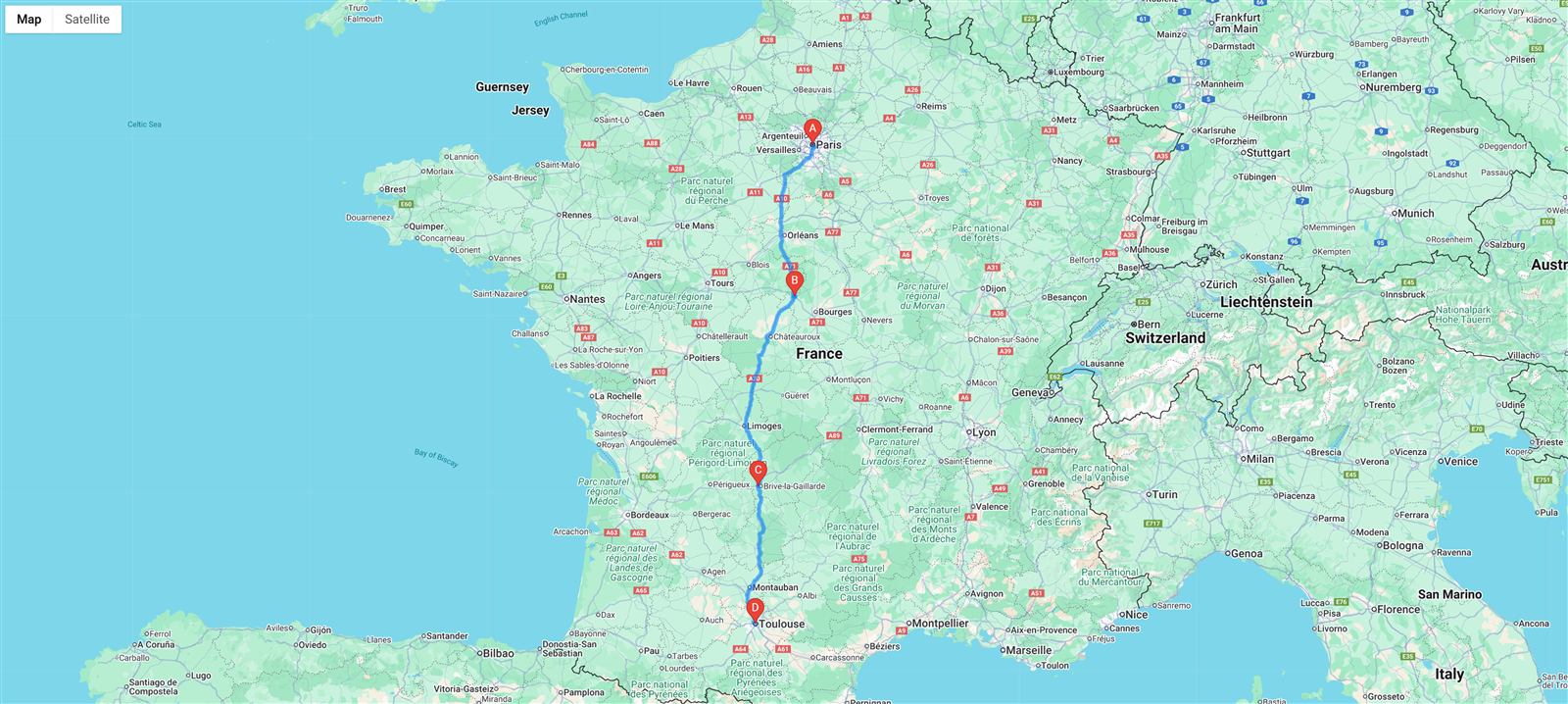}\\[-1mm]
\footnotesize (b) Paris--Toulouse
\end{minipage}\hfill
\begin{minipage}[t]{0.32\textwidth}
\centering
\includegraphics[width=\linewidth]{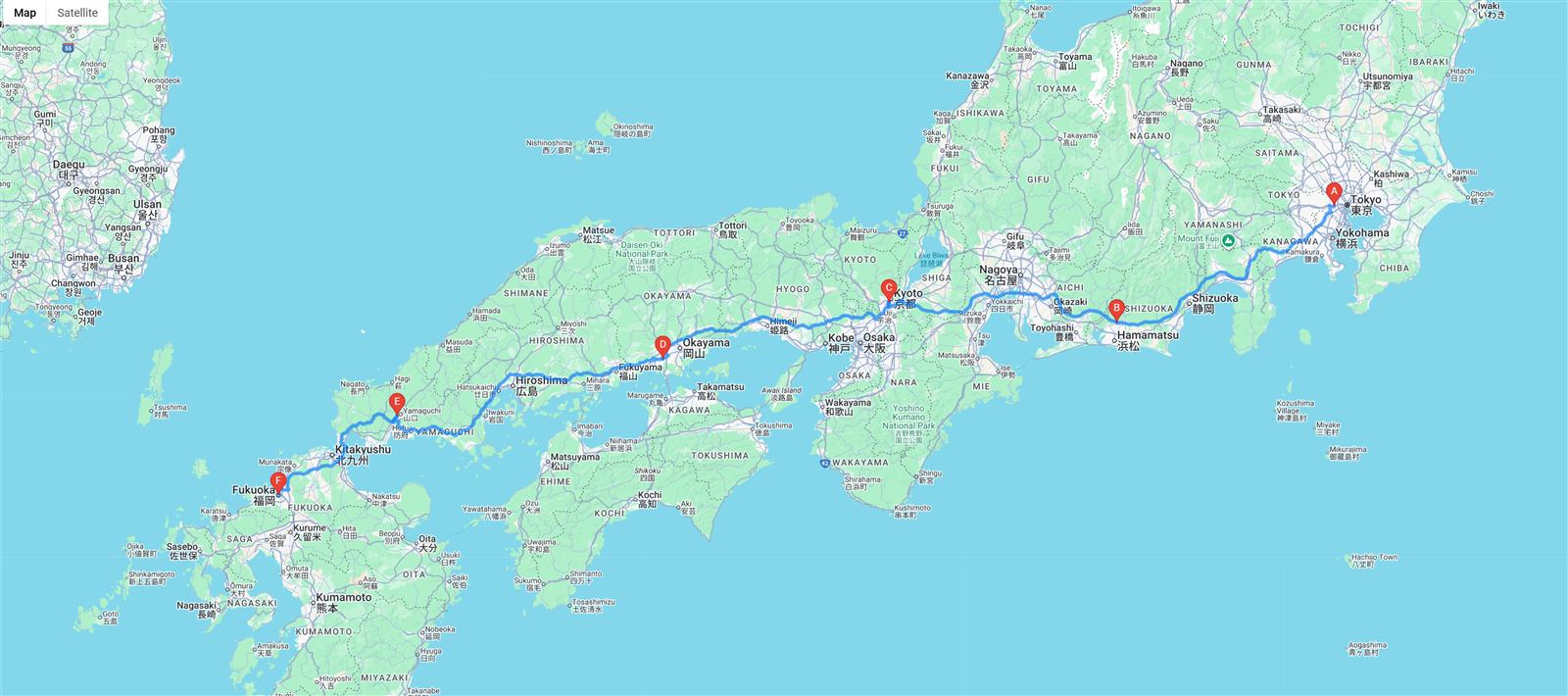}\\[-1mm]
\footnotesize (c) Tokyo--Fukuoka
\end{minipage}
\caption{Representative VEGA route visualizations for the three reported regions.}
\label{fig:route_visuals}
\end{figure*}

\subsection{Tesla Trip Planner Reference Case Study}\label{sec:tesla_case}

Tesla Trip Planner uses proprietary energy models and live data, so this comparison is qualitative rather than a controlled benchmark; VEGA matches Tesla's 20-stop plan but has a $2.12$\,h higher total trip time, including a $0.91$\,h charging-time gap explained by longer average dwell.

\subsection{International Zero-Shot Transfer}\label{sec:generalization}

Although trained only on U.S.\ routes, VEGA generalizes to international road networks.
Table~\ref{tab:generalization} summarizes zero-shot results across three regions using the same trained policy and the same U.S.-estimated PINO parameters.
Fig.~\ref{fig:route_visuals} shows representative VEGA routes for the three reported start/goal pairs.

\begin{table}[t]
\caption{Cross-Regional Generalization Summary}
\label{tab:generalization}
\centering
\footnotesize
\renewcommand{\arraystretch}{1.1}
\begin{tabular}{@{}lccc@{}}
\toprule
\textbf{Metric} & \textbf{US} & \textbf{France} & \textbf{Japan} \\
 & \footnotesize SF--NY & \footnotesize Paris--Toulouse & \footnotesize Tokyo--Fukuoka \\
\midrule
Distance (km) & 4{,}854.96 & 643.05 & 1{,}120.34 \\
Driving time (h) & 45.88 & 6.71 & 14.00 \\
Charging time (h) & 10.24 & 0.84 & 1.68 \\
Total trip time (h) & 56.12 & 7.55 & 15.68 \\
Charging stops & 20 & 2 & 4 \\
Energy (kWh) & 806.79 & 102.40 & 167.96 \\
Intensity (Wh/km) & 166.18 & 159.24 & 149.92 \\
Planning time (s) & $<2.0$ & $<1.0$ & $<1.0$ \\
\bottomrule
\end{tabular}
\vspace{1mm}

\footnotesize All routes computed with the same trained VEGA policy and PINO parameters estimated from U.S.\ driving data.
\end{table}

The agent successfully navigates all three routes without battery depletion, including the SF--NY case that is longer than the 3{,}000\,km maximum curriculum-training distance.
Energy intensity remains consistent across regions (149--166\,Wh/km), and the France (643\,km) and Japan (1{,}120\,km) routes require 2 and 4 charging stops, respectively.

\subsection{Design Choice Analysis}\label{sec:ablation}

\textbf{PINO vs.\ factory parameters:} inferred mass is +7.2\% and $C_d$ is +2.2\% versus reference values; using factory mass would underestimate long-route energy by about ${\sim}$7\%.
\textbf{Training strategy:} behavior cloning from A* stabilized full-graph exploration, while short-to-long curriculum progression (10$\to$300\,km, then expansion to 3{,}000\,km) reduced pathological looping near chargers.
\textbf{Long-horizon extrapolation:} the main SF--NY evaluation route (4{,}860\,km) is about 62\% longer than the maximum curriculum distance, so the reported benchmark already tests beyond the training horizon.
\textbf{Charging design:} the 80\% cap is motivated by Sec.~\ref{sec:data}, where 80\%$\to$100\% charging is time-inefficient.

\subsection{Discussion and Limitations}\label{sec:discussion}

VEGA demonstrates that coupling physics-informed parameter estimation with learned charge-aware routing is feasible on continental-scale road networks.
However, several simplifying assumptions warrant discussion.

\noindent\textbf{Flat terrain.}
We assume flat roads ($\alpha{=}0$), omitting the grade power term $mgv\sin\alpha$ from~\eqref{eq:pmech}.
On mountainous segments, this may underestimate energy consumption on uphills and overestimate on downhills.
Prior work shows that road grade materially affects EV energy and that downhill recovery is real but incomplete because regenerative braking efficiency is limited and condition-dependent~\cite{liu2017grade,bian2018regen_grade}.
Over long routes with mixed ascents/descents, some grade effects can partially cancel in aggregate, but local SoC errors can still be significant in steep terrain.
VEGA uses a 15\% low-SoC penalty threshold as a practical robustness buffer against this modeling error.
In our implementation, we intentionally keep altitude disabled for the controlled open-baseline comparison (VEGA, E-A*, Dijkstra+Greedy), where all three use the same flat-terrain energy model.
We already have altitude access via Google Maps Platform APIs (Roads/Elevation) and will add grade-aware routing in future work.

\noindent\textbf{Cruising at speed limits.}
We assume constant-speed cruising at the posted speed limit for each segment.
Real driving includes acceleration, deceleration, and traffic congestion, which can shift actual consumption.
Highway cruising is a reasonable first approximation for long-distance routing, but integration with real-time traffic speed data would improve segment energy estimates.

\noindent\textbf{Charger availability.}
We do not model charger queue times or station availability.
In practice, charger congestion can significantly affect total travel time, especially on popular corridors.

\noindent\textbf{Limitations of the study.}
Our main limitations are route coverage and sensitivity to parameter-estimation error.
We report three representative routes across three regions (643--4{,}855\,km) due to page limits.
Additional start/goal pairs were explored during development, but they are not part of the reported benchmark here; a broader benchmark would better quantify variance across route types.
The routing agent uses PINO-estimated parameters as input to the energy model, so if PINO estimates are inaccurate (e.g., due to unusual driving conditions in the calibration window), the resulting energy predictions can be biased.
The low-SoC penalty in the reward is intended to improve robustness against this modeling error, potentially at the cost of longer charging sessions.
From~\eqref{eq:soc_discharge}, an energy prediction error $\delta E$ (kWh) induces an SoC error $\delta b = 100\,\delta E/B$ (\%); with $B{=}75$\,kWh (Sec.~\ref{sec:data}), a 1\,kWh error corresponds to about 1.33 SoC points.

\section{CONCLUSION}\label{sec:conclusion}

We presented VEGA, a system that integrates physics-informed neural operator (PINO)-based vehicle parameter estimation with a PPO-trained charge-aware routing agent for electric vehicles.
VEGA operates on real continental-scale road networks with charger infrastructure, using only onboard speed and acceleration logs to adapt its energy model to the current vehicle condition.

On the transcontinental San Francisco--to--New York benchmark, VEGA produces a feasible 20-stop plan with 56.12\,h total trip time and minimum SoC 11.41\%.
The remaining gap to Energy-aware A* is dominated by charging dwell time rather than route distance or driving time, while inference remains $>$20$\times$ faster.
The learned policy generalizes to unseen road networks in France and Japan without retraining.

Future work includes incorporating road grade and real-time traffic data into the energy model, validating across multiple vehicle models, extending the evaluation to a large-scale benchmark with diverse route characteristics, and deploying VEGA for real-time onboard replanning on production EVs.

\bibliographystyle{IEEEtran}
\bibliography{references}

@article{li2023sdv,
  title={Software-defined vehicles: Architecture, challenges, and perspectives},
  author={Li, L. and others},
  journal={IEEE Trans. Intell. Veh.},
  year={2023},
  volume={8},
  number={6},
  pages={3803--3822}
}

@article{miri2021ev_energy,
  title={Electric vehicle energy consumption estimation for routing applications},
  author={Miri, I. and Fotouhi, A. and Ewin, N.},
  journal={J. Clean. Prod.},
  year={2021},
  volume={326},
  pages={129--246}
}

@article{basso2019ev_energy_survey,
  title={Energy consumption estimation integrated into the electric vehicle routing problem},
  author={Basso, R. and Kulcs{\'a}r, B. and Egardt, B. and Lindroth, P. and Sanchez-Diaz, I.},
  journal={Transp. Res. Part D},
  year={2019},
  volume={69},
  pages={141--167}
}

@article{montoya2017evrp,
  title={The electric vehicle routing problem with nonlinear charging function},
  author={Montoya, A. and Gu{\'e}ret, C. and Mendoza, J. E. and Villegas, J. G.},
  journal={Transp. Sci.},
  year={2017},
  volume={51},
  number={4},
  pages={1234--1249}
}

@inproceedings{schneider2014evrptw,
  title={The electric vehicle routing problem with time windows and recharging stations},
  author={Schneider, M. and Stenger, A. and Goeke, D.},
  booktitle={Transp. Sci.},
  year={2014},
  volume={48},
  pages={500--520}
}

@article{li2021fno,
  title={Fourier neural operator for parametric partial differential equations},
  author={Li, Z. and Kovachki, N. and Azizzadenesheli, K. and Liu, B. and Bhatt, K. and Stuart, A. and Anandkumar, A.},
  journal={arXiv preprint arXiv:2010.08895},
  year={2021}
}

@inproceedings{lim2024pino_ev,
  title={Physics-informed neural operator for real-time {EV} parameter estimation},
  author={Lim, H. and Im, M. and Boyack, J. and Choi, J. B.},
  booktitle={Proc. IEEE Int. Conf. Intell. Transp. Syst. (ITSC)},
  year={2024},
  pages={1--8}
}

@article{schulman2017ppo,
  title={Proximal policy optimization algorithms},
  author={Schulman, J. and Wolski, F. and Dhariwal, P. and Radford, A. and Klimov, O.},
  journal={arXiv preprint arXiv:1707.06347},
  year={2017}
}

@article{wang2023ev_fno,
  title={Physics-informed approaches for electric vehicle battery power estimation from speed alone},
  author={Wang, X. and others},
  journal={Appl. Energy},
  year={2023},
  volume={340},
  pages={121009}
}

@article{raissi2019pinn,
  title={Physics-informed neural networks: A deep learning framework for solving forward and inverse problems involving nonlinear partial differential equations},
  author={Raissi, M. and Perdikaris, P. and Karniadakis, G. E.},
  journal={J. Comput. Phys.},
  year={2019},
  volume={378},
  pages={686--707}
}

@inproceedings{kool2019vrp_attention,
  title={Attention, learn to solve routing problems!},
  author={Kool, W. and van Hoof, H. and Welling, M.},
  booktitle={Int. Conf. Learn. Represent. (ICLR)},
  year={2019}
}

@inproceedings{nazari2018vrp_rl,
  title={Reinforcement learning for solving the vehicle routing problem},
  author={Nazari, M. and Oroojlooy, A. and Snyder, L. V. and Tak{\'a}c, M.},
  booktitle={Advances in Neural Inf. Process. Syst. (NeurIPS)},
  year={2018},
  pages={9839--9849}
}

@article{chen2022ev_path_planning,
  title={Reinforcement learning for electric vehicle path planning with charging considerations},
  author={Chen, S. and others},
  journal={IEEE Trans. Intell. Transp. Syst.},
  year={2022},
  volume={23},
  number={8},
  pages={12744--12755}
}

@article{joshi2022gnn_routing,
  title={Learning heuristics for combinatorial optimization on graphs using deep reinforcement learning},
  author={Joshi, C. and Laurent, T. and Bresson, X.},
  journal={arXiv preprint arXiv:1903.04864},
  year={2022}
}

@article{swazinna2021overcoming,
  title={Overcoming model bias for robust offline deep reinforcement learning},
  author={Swazinna, P. and Udluft, S. and Runkler, T.},
  journal={Eng. Appl. Artif. Intell.},
  year={2021},
  volume={104},
  pages={104366}
}

@article{bengio2021ml4co,
  title={Machine learning for combinatorial optimization: A methodological tour d'horizon},
  author={Bengio, Y. and Lodi, A. and Prouvost, A.},
  journal={European J. Oper. Res.},
  year={2021},
  volume={290},
  number={2},
  pages={405--421}
}

@inproceedings{kwon2020pomo,
  title={{POMO}: Policy optimization with multiple optima for reinforcement learning},
  author={Kwon, Y. D. and Choo, J. and Kim, B. and Yoon, I. and Gwon, Y. and Min, S.},
  booktitle={Advances in Neural Inf. Process. Syst. (NeurIPS)},
  year={2020},
  pages={21188--21198}
}

@article{schulze2020gae,
  title={High-dimensional continuous control using generalized advantage estimation},
  author={Schulman, J. and Moritz, P. and Levine, S. and Jordan, M. and Abbeel, P.},
  journal={arXiv preprint arXiv:1506.02438},
  year={2016}
}

@article{bengio2020curriculum,
  title={Curriculum learning for reinforcement learning domains: A framework and survey},
  author={Narvekar, S. and Peng, B. and Leonetti, M. and Sinapov, J. and Taylor, M. E. and Stone, P.},
  journal={ACM Comput. Surv.},
  year={2020},
  volume={54},
  number={7},
  pages={1--36}
}

@article{fiori2016ev_power,
  title={Power-based electric vehicle energy consumption model: Model development and validation},
  author={Fiori, C. and Ahn, K. and Rakha, H. A.},
  journal={Appl. Energy},
  year={2016},
  volume={168},
  pages={257--268}
}

@article{liu2017grade,
  title={Impact of road gradient on energy consumption of electric vehicles},
  author={Liu, Kai and Yamamoto, Toshiyuki and Morikawa, Takayuki},
  journal={Transportation Research Part D: Transport and Environment},
  year={2017},
  volume={54},
  pages={74--81},
  doi={10.1016/j.trd.2017.05.005}
}

@article{bian2018regen_grade,
  title={Effect of road gradient on regenerative braking energy in a pure electric vehicle},
  author={Bian, Jian and Qiu, Bo},
  journal={Proceedings of the Institution of Mechanical Engineers, Part D: Journal of Automobile Engineering},
  year={2018},
  volume={232},
  number={13},
  pages={1736--1746},
  doi={10.1177/0954407017735020}
}

@article{hart1968astar,
  title={A formal basis for the heuristic determination of minimum cost paths},
  author={Hart, P. E. and Nilsson, N. J. and Raphael, B.},
  journal={IEEE Trans. Syst. Sci. Cybern.},
  year={1968},
  volume={4},
  number={2},
  pages={100--107}
}

\end{document}